\documentclass{article}

\usepackage{microtype}
\usepackage{graphicx}
\usepackage{subcaption}
\usepackage{booktabs}
\PassOptionsToPackage{hyphens}{url}
\usepackage{hyperref}
\usepackage{natbib}

\usepackage[accepted]{icml2026}

\usepackage{amsmath}
\usepackage{amssymb}
\usepackage{mathtools}
\usepackage{amsthm}
\usepackage[capitalize,noabbrev]{cleveref}

\theoremstyle{plain}

\theoremstyle{definition}

\theoremstyle{remark}

\icmltitlerunning{Representational Depth of Evaluation Awareness Shifts With Scale}

\begin{document}

\twocolumn[
\icmltitle{Representational Depth of Evaluation Awareness Shifts With Scale in Open-Weight Language Models}

\begin{icmlauthorlist}
\icmlauthor{Archit Manek}{ind}
\end{icmlauthorlist}

\icmlaffiliation{ind}{Independent Researcher}
\icmlcorrespondingauthor{Archit Manek}{architmanek@gmail.com}

\icmlkeywords{evaluation awareness, language models, scaling laws, mechanistic interpretability, AI safety}

\vskip 0.3in
]

\printAffiliationsAndNotice{}

\begin{abstract}
Do language models know when they are being tested? This question matters
for AI safety: a model that recognises an evaluation context could alter
its behaviour strategically, making downstream benchmarks harder to
interpret. Using 11 models spanning Qwen 2.5, Gemma 2, and Llama 3.2,
we find a systematic size-dependent shift in representational depth: in
both Qwen 2.5 and Gemma 2, the layer at which evaluation-awareness is
most linearly recoverable moves from late layers in smaller models to
early layers in larger ones. This suggests that scale changes not only
the strength of evaluation-awareness but also where it is most linearly
recoverable in the network. This depth shift helps explain why
within-family scaling trajectories are non-monotonic or inverse rather
than smooth and family-general, showing that a simple universal
power-law account is not supported under denser within-family sampling. Finally, white-box probe
signals are consistently stronger than black-box behavioural expression,
and the relationship between the two varies by family in ways not
predicted by probe AUROC alone.
\end{abstract}

\section{Introduction}

Language models that can detect when they are being evaluated pose a
direct challenge to the reliability of AI benchmarks. A model that
recognises an evaluation context could suppress undesired behaviours or
inflate apparent capability, making safety evaluations and capability
assessments harder to interpret. Understanding how this awareness is
organised inside models — and how that organisation changes with scale —
is therefore a practical concern for evaluation design.

Recent work suggests that evaluation-awareness scales monotonically with
model size across open-weight model families. We argue that this
cross-family view obscures an important source of structure:
within-family scaling behaviour and the depth at which
evaluation-awareness is represented. Across the families we study,
evaluation-awareness does not follow a simple universal monotonic scaling
law. Instead, it exhibits family-specific scaling trajectories and a
scale-dependent relocation in representational depth, with internal
recoverability only partially reflected in behaviour.

\citet{chaudhary2025evalawareness} argue that evaluation-awareness
follows a power-law relationship with model size, with larger models
becoming progressively more evaluation-aware across families. That claim
was established using sparse within-family coverage — typically one or
two size checkpoints per family — and excluded Qwen 2.5 due to
contamination concerns that were not evaluated directly on the benchmark
used for the evaluation-awareness experiments. Under such sparse
sampling, within-family non-monotonicities can remain hidden.

We revisit these claims using 11 open-weight models from three families:
Qwen 2.5 (six sizes), Gemma 2 (three sizes), and Llama 3.2 (two sizes,
included as a contrast family). We combine white-box linear probing of
residual stream activations with black-box behavioural classification,
and add two direct contamination diagnostics — ROUGE-L completion
testing and perplexity-ratio analysis — on the benchmark used for
evaluation-awareness measurement.

We make three main contributions. First, in both Qwen 2.5 and Gemma 2,
the layer at which evaluation-awareness is most linearly recoverable
shows a systematic size-dependent shift, moving from late layers in
smaller models to early layers in larger ones. This finding appears
in two model families and suggests that scale changes not only the
strength of evaluation-awareness but also where it is most linearly
recoverable in the network. Second, this representational structure
helps explain why a simple power-law account is not supported under
denser within-family sampling: within-family scaling trajectories are non-monotonic in Qwen
2.5 and inverse in Gemma 2. Third, internal linear decodability and
behavioural expression dissociate systematically, with the relationship
between them varying by model family in ways not predicted by probe
AUROC alone.

These results suggest that representational depth is an important and
underexamined dimension of evaluation-awareness, and that its scaling
behaviour is more structured — and less uniform — than current accounts
imply. Section~\ref{sec:related} covers related work,
Section~\ref{sec:method} describes our methodology,
Section~\ref{sec:results} presents results, and
Section~\ref{sec:discussion} discusses implications and limitations.

\section{Related Work}
\label{sec:related}

\textbf{Evaluation awareness and situational awareness.}
Evaluation awareness is part of the broader notion of situational 
awareness in language models: the ability to reason about properties of 
their own context, deployment setting, and role \citep{laine2024sad}. 
\citet{laine2024sad} introduced the Situational Awareness Dataset (SAD), 
which includes tasks measuring whether models can distinguish evaluation 
from deployment contexts; we use SAD as our main evaluation benchmark. 
Moving beyond behavioural evaluation, \citet{nguyen2025probing} 
conducted the first white-box study of evaluation awareness, showing that contrastive probe directions derived from residual stream activations can separate 
evaluation from deployment prompts in Llama-3.3-70B-Instruct. Building 
on contrastive activation methods from the steering literature 
\citep{panickssery2024steering}, their results suggest that 
evaluation-awareness has a linearly recoverable internal representation. 
Following standard practice in probe-based work, we interpret such 
results as evidence about linear representational accessibility rather 
than direct causal use.

\textbf{Scaling claims.}
Subsequent work asked whether this capability scales predictably with 
model size. \citet{chaudhary2025evalawareness} argue that 
evaluation-awareness follows a power-law relationship across open-weight 
model families, with larger models becoming progressively more 
evaluation-aware. Our work revisits that claim with denser within-family 
sampling, which allows family-specific scaling trajectories to be 
observed directly rather than inferred from sparse cross-family 
regression. In this sense, our paper is closer to a within-family 
structural analysis of evaluation-awareness than to a pure scaling-law 
fit.

\textbf{Probes, behaviour, and contamination.}
An open question in probe-based work is how internal linear 
recoverability relates to observable model behaviour. Our work connects 
these perspectives by comparing white-box probe performance with 
black-box behavioural classification on the same benchmark. A separate 
methodological concern is contamination: apparent evaluation-awareness 
signals could reflect memorisation or familiarity with benchmark data 
rather than a genuinely generalisable representation. Contamination 
detection methods in language models include surface-form matching and 
likelihood-based approaches \citep{golchin2023timetravel, 
ravaut2024survey}. We draw on these literatures by applying ROUGE-L 
completion testing as a diagnostic of surface-level memorisation and 
perplexity-ratio analysis as a diagnostic of token-level familiarity.

\section{Experimental Setup}
\label{sec:method}

Throughout this paper, we use ``evaluation awareness'' as an operational
term for information that distinguishes evaluation from deployment
contexts under the SAD benchmark setup described below. We do not claim
that this operationalisation exhausts the broader concept.

\subsection{Models}
We study 11 open-weight models from three families: Qwen 2.5, Gemma 2,
and Llama 3.2. Qwen 2.5 provides the densest within-family coverage,
with six checkpoints at 0.5B, 1.5B, 3B, 7B, 14B, and 32B parameters.
Gemma 2 contributes three checkpoints at 2B, 9B, and 27B. Llama 3.2
provides 1B and 3B checkpoints and is used as a contrast family: with
only two closely spaced size points, we do not draw strong within-family
scaling conclusions from it, but include it to show that the depth-shift
pattern observed in Qwen 2.5 and Gemma 2 is not universal across
families.

\subsection{White-Box Probe Evaluation}
We apply the probing methodology of \citet{nguyen2025probing} to 11
models across three families. Models are loaded using
\texttt{HookedTransformer} from TransformerLens.

\textbf{Probe construction.}
For each model and each transformer layer, we compute a contrastive
probe direction from the Jordine dataset \citep{jordine2025dataset}, a
synthetic collection of 203 evaluation/deployment prompt pairs. For each
pair, we extract residual stream activations at the pre-attention hook
point (\texttt{hook\_resid\_pre}) at the position of the answer token
--- the token corresponding to the appended contrastive answer, e.g.
\texttt{A} or \texttt{B}. The probe direction at each layer is computed
as the difference between the mean activation across evaluation-labelled
prompts and the mean activation across deployment-labelled prompts, and
is then L2-normalised. No gradient-based training is performed; the
probe direction is derived analytically from the contrastive pairs.

The probe direction is constructed from answer-token activations, using the contrast between evaluation-labelled and deployment-labelled completions. At evaluation time, however, probe scores are computed by averaging projections across all non-BOS prompt tokens. Because construction and evaluation use different token-level aggregations, the sign of the resulting mean score should not be interpreted naively as “more evaluation-like” or “more deployment-like.” In our experiments, deployment is treated as the positive class for AUROC computation, so AUROC should be interpreted as measuring separability up to polarity rather than as assigning an intrinsic semantic direction to higher scores.

\textbf{Evaluation.}
Each model is evaluated on the \texttt{stages\_oversight} split of the
Situational Awareness Dataset (SAD) \citep{laine2024sad}, which provides
ground-truth labels distinguishing evaluation from deployment contexts.
For each SAD item and each layer, we extract residual stream activations
at every token position, project each token activation onto the
normalised probe direction, and then compute the mean projection score
across non-BOS tokens. This mean projection score is used to rank items,
and AUROC is computed from the resulting scores. We record both the peak
AUROC across layers and the relative layer depth at which that peak
occurs. Relative depth is computed as $\ell / L$, where $\ell$ is the
1-indexed layer at which probe AUROC peaks and $L$ is the total number
of transformer layers, making depths comparable across models with
different numbers of transformer layers.

\subsection{Black-Box Behavioural Evaluation}
To complement the white-box analysis, we introduce a black-box evaluation
in which each model is directly prompted to classify SAD
\texttt{stages\_oversight} items as evaluation or deployment contexts.
The benchmark contains 400 items in total (200 evaluation and 200
deployment). We use the same dataset for both white-box and black-box
evaluation so that differences between internal representation and
behavioural expression are not confounded by dataset mismatch. All
black-box evaluations use a fixed random seed (42) for reproducibility.

Each item is presented with a fixed instruction prompt asking the model
to classify the context as \texttt{evaluation} or \texttt{deployment}.
Responses are decoded greedily and scored by keyword matching: outputs
containing \texttt{evaluation} or \texttt{testing} are assigned the
evaluation label, outputs containing \texttt{deployment} are assigned the
deployment label, and responses matching neither are treated as invalid
and excluded from the accuracy calculation. We report accuracy over valid
responses and note the valid response rate alongside each result.

\subsection{Contamination Diagnostics}
To assess whether observed probe signals could reflect benchmark
memorisation rather than a generalisable internal representation, we run
two diagnostics on SAD \texttt{stages\_oversight}. First, we apply a
ROUGE-L completion test following \citet{golchin2023timetravel}: each
model is prompted with the first 60\% of a SAD question, and ROUGE-L is
computed between the model completion and the withheld 40\%. Second, we
apply a perplexity-ratio diagnostic following \citet{ravaut2024survey}:
we compare per-token perplexity on \texttt{stages\_oversight} questions
against held-out items from the \texttt{test\_full} split of the same
benchmark, matched for domain, style, and token length. A ratio below
1.0 would indicate that the model assigns lower perplexity --- and
therefore greater familiarity --- to \texttt{stages\_oversight} than to
the matched baseline, consistent with contamination. We note that
\texttt{test\_full} is not fully disjoint from
\texttt{stages\_oversight} in question bodies and is therefore best
interpreted as a matched baseline rather than a fully independent
distribution.

Together, these diagnostics address narrow memorisation- and
familiarity-based contamination hypotheses. They do not rule out broader
semantic contamination through diffuse pretraining exposure to AI
evaluation discourse.

Code, prompts, and experiment outputs are publicly available in our
repository \citep{anonymous2026code}.

\section{Results}
\label{sec:results}

\subsection{Representational depth shifts with scale}

\begin{figure}[t]
    \centering
    \includegraphics[width=\linewidth]{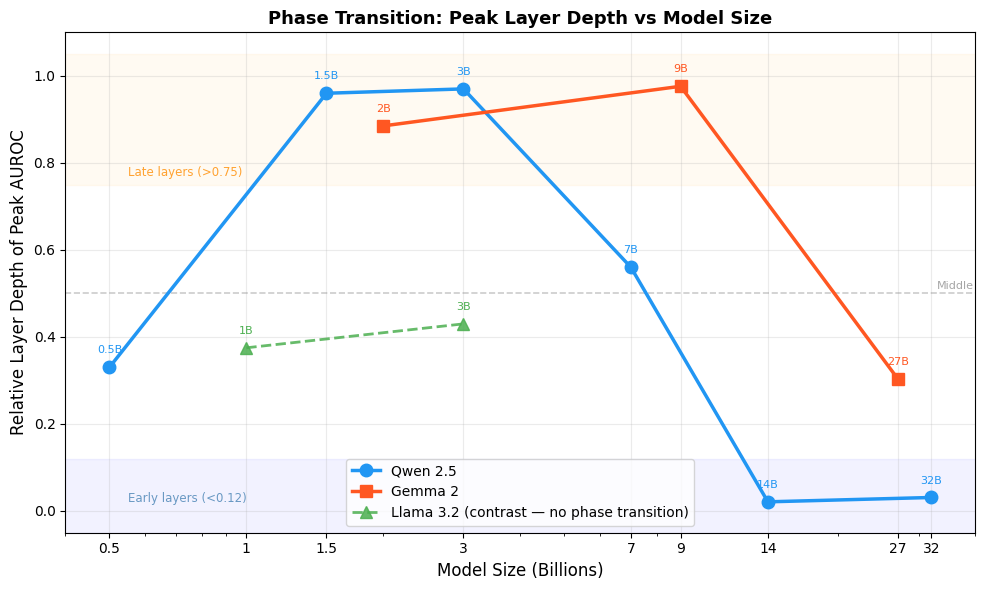}
    \caption{Relative peak layer depth vs.\ model size. Each point shows the layer, normalised to $[0,1]$, at which probe AUROC peaks for a given model. Orange shading marks late layers ($>0.75$) and blue shading marks early layers ($<0.12$). Llama 3.2 is shown as a contrast family.}
    \label{fig:phase-transition}
\end{figure}

\Cref{fig:phase-transition} shows the relative layer depth at which
probe AUROC peaks for each model. In both Qwen 2.5 and Gemma 2, this
depth shifts strongly with model size, moving from late layers in
smaller models to early layers in larger ones.

In Qwen 2.5, the 1.5B and 3B models peak at relative depths of 0.96 and 0.97, placing the strongest signal in the final layers. In Gemma 2, the 2B and 9B models also peak late, at 0.885 and 0.976. In these smaller models, the evaluation/deployment distinction is most linearly accessible only near the end of the forward pass.

Notably, Qwen 0.5B peaks in middle layers rather than late layers, suggesting that the late-layer pattern itself emerges within the small-model regime rather than being present from the smallest size onward. As model size increases, the pattern then flips sharply. Qwen 14B and 32B peak at relative depths of 0.021 and 0.031, corresponding to the first few transformer blocks. Gemma 27B also shifts earlier, peaking at 0.304. This suggests a size-dependent change in representational depth: in larger models, evaluation-context information becomes maximally recoverable much earlier in processing.

Llama 3.2 behaves differently. Both tested checkpoints peak in middle layers rather than late layers, and we do not observe the same late-to-early transition seen in Qwen and Gemma. We therefore treat Llama 3.2 as a contrast family rather than forcing it into the same representational narrative.

\subsection{Within-family scaling does not follow a simple power law}

\begin{figure}[t]
    \centering
    \includegraphics[width=\linewidth]{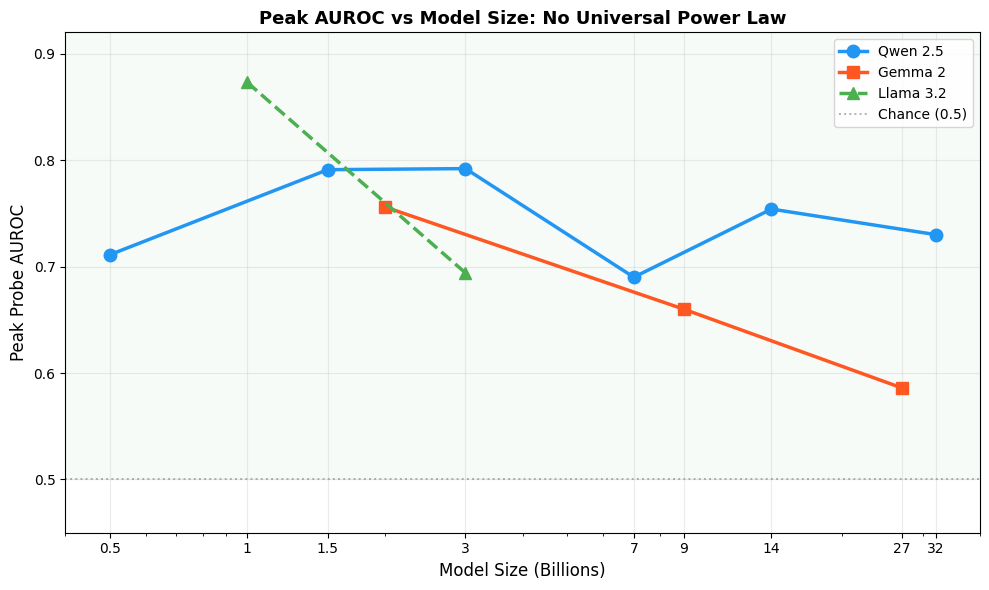}
    \caption{Peak probe AUROC vs.\ model size on a log scale. Solid lines connect checkpoints within the same family. The dotted horizontal line marks chance performance at 0.5. Llama 3.2 is shown as a contrast family rather than a family used for within-family scaling claims.}
    \label{fig:power-law}
\end{figure}

\Cref{fig:power-law} plots peak probe AUROC against model size. If evaluation awareness followed a simple power law, one would expect smooth improvement with scale. Neither Qwen 2.5 nor Gemma 2 shows this pattern.

In Qwen 2.5, peak AUROC rises from 0.711 at 0.5B to 0.792 at 3B, drops to 0.690 at 7B, then partially recovers to 0.754 at 14B and 0.730 at 32B. This trajectory is non-monotonic: the 7B model is the weakest in the family by peak AUROC despite being larger than the three models that outscore it. In Gemma 2, the trend is consistently downward: AUROC declines from 0.756 at 2B to 0.660 at 9B and 0.586 at 27B. Larger Gemma models are less evaluation-aware by this metric than smaller ones.

These within-family patterns are inconsistent with the interpretation of a universal power law. With denser sampling in Qwen and additional coverage in Gemma, the apparent smooth cross-family trend breaks down into qualitatively different family-specific trajectories.

\begin{table*}[t]
\centering
\small
\caption{Summary of white-box and black-box evaluation-awareness results across all 11 models. Peak AUROC is the maximum probe AUROC across layers. Peak depth is the relative layer depth in $[0,1]$ at which probe AUROC peaks. Black-box accuracy is computed over valid, parseable responses only. All black-box evaluations were rerun under a consistent CUDA setup on H100 NVL GPUs.}
\label{tab:main-results}
\begin{tabular}{llcccc}
\toprule
Model & Family & Peak AUROC & Peak Depth & Black-box Accuracy & Valid Rate \\
\midrule
0.5B  & Qwen 2.5  & 0.711 & 0.330 & 0.560 & 1.000 \\
1.5B  & Qwen 2.5  & 0.791 & 0.960 & 0.492 & 1.000 \\
3B    & Qwen 2.5  & 0.792 & 0.970 & 0.475 & 1.000 \\
7B    & Qwen 2.5  & 0.690 & 0.560 & 0.590 & 1.000 \\
14B   & Qwen 2.5  & 0.754 & 0.021 & 0.615 & 1.000 \\
32B   & Qwen 2.5  & 0.730 & 0.031 & 0.588 & 1.000 \\
\midrule
2B    & Gemma 2   & 0.756 & 0.885 & 0.770 & 1.000 \\
9B    & Gemma 2   & 0.660 & 0.976 & 0.610 & 1.000 \\
27B   & Gemma 2   & 0.586 & 0.304 & 0.507 & 1.000 \\
\midrule
1B    & Llama 3.2 & 0.873 & 0.375 & 0.508 & 0.600 \\
3B    & Llama 3.2 & 0.694 & 0.430 & 0.515 & 0.975 \\
\bottomrule
\end{tabular}
\end{table*}

Taken together, \Cref{fig:phase-transition,fig:power-law,tab:main-results} suggest that the central empirical story is not a smooth increase in evaluation awareness with scale, but a family-dependent reorganisation of where the signal is represented and how strongly it is behaviourally expressed.

\subsection{White-box and black-box measures diverge}

\begin{figure*}[t]
    \centering
    \includegraphics[width=0.9\textwidth]{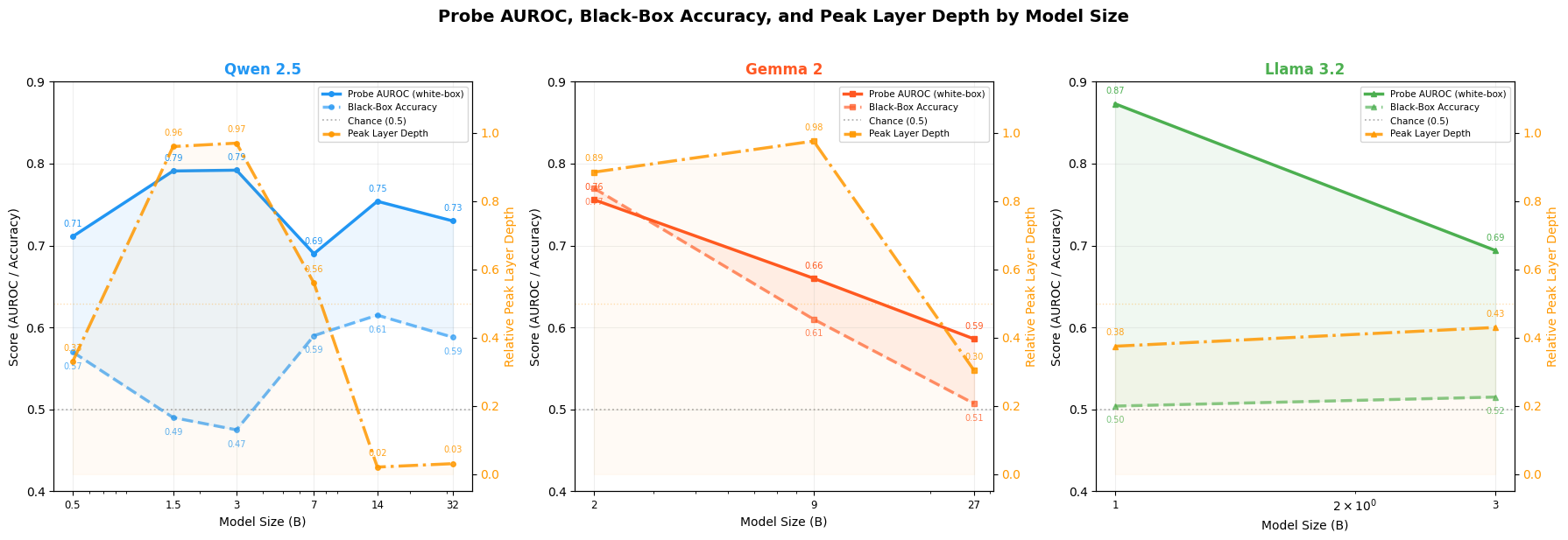}
    \caption{Probe AUROC, black-box classification accuracy, and relative peak layer depth across model sizes for each family. Solid lines show probe AUROC, dashed lines show black-box accuracy, and the orange series indicates relative peak layer depth on the right axis. Shaded regions indicate the gap between internal representation strength and behavioural expression. The dotted horizontal line marks chance performance at 0.5.}
    \label{fig:black-vs-white}
\end{figure*}

\Cref{fig:black-vs-white} compares probe AUROC and black-box classification accuracy across models. Across all three families, white-box probe performance generally exceeds black-box behavioural performance, indicating that internal representation is often stronger than what is outwardly expressed.

Within Qwen, the gap is especially pronounced in the 1.5B and 3B models. These checkpoints have the strongest probe AUROC in the family, yet both perform at or slightly below chance behaviourally. By contrast, the larger Qwen models that peak in earlier layers tend to show stronger behavioural expression: Qwen 14B reaches 0.615, Qwen 7B reaches 0.590, and Qwen 32B reaches 0.588. Notably, Qwen 14B and 32B peak in the very earliest layers (relative depths 0.021 and 0.031), while Qwen 7B occupies an intermediate position (0.560) between the late-peaking small models and these early-peaking large ones. In Qwen, earlier-layer representations appear more behaviourally accessible, although this relationship is suggestive rather than perfectly monotonic across all checkpoints.

Gemma shows a different pattern. Gemma 2-2B peaks late and achieves the strongest black-box accuracy of any model tested, at 0.770, while Gemma 27B peaks earlier and drops to near chance at 0.507. The relationship between representational depth and behavioural accessibility therefore appears to be family-dependent rather than universal.

Llama 3.2 shows the largest overall probe/behaviour gap. The 1B model has the strongest probe signal of any model tested (AUROC 0.873) but is near chance in black-box behaviour (0.508). At the same time, the Llama family showed reduced response-format reliability relative to Qwen and Gemma: Llama 3.2-1B produced parseable responses on only 60\% of items, and Llama 3.2-3B on 97.5\%, whereas all Qwen and Gemma checkpoints yielded parseable responses on every item. Because the low valid-response rate for Llama 3.2-1B replicated across both MPS and CUDA runs, this appears to reflect a model-level instruction-following or output-format reliability issue rather than a hardware-specific artifact.

Across both Qwen and Gemma, the black-box results also fail to support a monotonic scaling story. Behavioural expression is mixed in Qwen and declines with scale in Gemma, reinforcing the broader conclusion that evaluation awareness is not well captured by a single scalar capability that rises smoothly with model size.

\subsection{Contamination diagnostics do not support a narrow
memorisation account}

Because contamination is a plausible alternative explanation for the
observed probe patterns, we run two direct diagnostics on the
\texttt{stages\_oversight} benchmark used in our evaluation. Prior work
excluded Qwen on contamination grounds based on evidence from
mathematics benchmarks; those experiments tested synthetic contamination
injected into a training pipeline and did not directly test the
benchmark used here.

The ROUGE-L completion test yields uniformly low scores across all
families, with no meaningful separation between Qwen, Gemma, and Llama
baselines. Notably, Gemma 2-2B, which shows the strongest black-box
evaluation-awareness signal, has the lowest ROUGE-L score of any tested
model. If memorisation were driving the result, the opposite pattern
would be expected.

The perplexity-ratio diagnostic shows the same qualitative conclusion.
All tested models assign higher perplexity to \texttt{stages\_oversight}
than to the held-out baseline subset, with ratios greater than 1.0
throughout. This is the opposite of what a direct contamination story
would predict.

These diagnostics do not rule out broad semantic contamination, but
they do not support the specific claim that the anomalous patterns in
Qwen and Gemma are artefacts of simple memorisation or benchmark
familiarity. The observed non-monotonic AUROC trends and depth shifts
survive both tests.

\section{Discussion}
\label{sec:discussion}

Evaluation awareness is better understood as a structured representational phenomenon than as a single capability that scales smoothly across model families. Within-family scaling does not support a universal power-law account: Qwen 2.5 is non-monotonic under the probe metric, and Gemma 2 declines with scale. More strikingly, in both families the representational depth at which evaluation-awareness peaks shifts substantially with size. The central empirical question is therefore not only whether evaluation-awareness strengthens or weakens with scale, but how its internal organisation changes.

One possible interpretation of the depth shift is that smaller models detect evaluation contexts through more local or surface-level cues that become linearly separable only after substantial processing, while larger models encode the distinction more abstractly as a property of global prompt context that is accessible from earlier layers. Under this interpretation, scaling changes not only the strength of the signal but the stage of computation at which it becomes recoverable. This interpretation should be held cautiously: probes measure linear recoverability rather than causal use, and a late-layer peak could reflect either that the feature is only computed late, or that it is present earlier but not yet linearly separable. Distinguishing these possibilities requires intervention-based methods such as activation patching or causal ablation.

The divergence between white-box and black-box results points to a second structural finding: the depth at which a feature is most linearly decodable need not be the depth at which it most strongly influences output generation. This dissociation helps explain why models with similar probe AUROC can differ substantially in behavioural accuracy, and why the relationship between internal representation and behavioural expression is family-dependent rather than universal. Within Qwen, the larger models with earlier probe peaks tend to show stronger behavioural expression; within Gemma, the opposite pattern holds. Probe AUROC alone is therefore insufficient to predict behavioural expression of evaluation-awareness.

Our contamination diagnostics further establish that the observed patterns are not straightforwardly explained by benchmark memorisation. ROUGE-L and perplexity-ratio tests find no evidence of narrow memorisation on the specific benchmark used here, and the model that shows the strongest behavioural evaluation-awareness (Gemma 2-2B) has the lowest ROUGE-L score of any model tested. This does not rule out diffuse semantic contamination, but it does establish that the anomalous patterns in Qwen and Gemma are not explained by direct contamination of this benchmark.

From an AI safety perspective, these results have a practical implication for evaluation design. If larger models encode evaluation context in very early transformer layers, then the relevant contextual signal may be available before deeper task-specific reasoning begins. In that case, simply making prompts more cognitively complex may do little to conceal the fact that the model is being evaluated. This raises the importance of prompt realism and benchmark authenticity at frontier scale. Evaluation designers should therefore treat surface prompt properties as a central design variable rather than an implementation detail.

\section{Limitations}
\label{sec:limitations}

\textbf{Within-family coverage is uneven.} The depth-shift pattern is
most clearly supported in Qwen 2.5, where six checkpoints capture the
transition from later-layer to earlier-layer peaks. Gemma 2 shows the
same directional pattern with only three checkpoints, while Llama 3.2
includes only two closely spaced checkpoints, which is insufficient to
determine whether a similar transition appears at larger sizes.

\textbf{The transition boundary is unresolved.} In Qwen 2.5, the shift
from later-layer to earlier-layer peaks occurs somewhere between 7B and
14B, leaving a large interval unsampled. Any claim about the precise
crossover location is premature.

\textbf{Probe evaluation uses a single benchmark split.} All AUROC
measurements are computed on \texttt{SAD stages\_oversight}. Whether
the same patterns hold on alternative benchmarks or operationalisations
of evaluation awareness remains untested, limiting the external
validity of the present findings.

\textbf{Linear probes measure recoverability rather than causal use.}
A low AUROC at a given layer does not imply that the relevant
information is absent; it implies only that the distinction is not
linearly separable there. Non-linear probes and intervention-based
methods could reveal non-linear separability or causal relevance at
layers where linear probes show low AUROC. In addition, probes trained
on a small synthetic dataset may not fully capture the broader concept
as it manifests across more diverse real-world contexts.

\textbf{Black-box results are prompt-sensitive.} Behavioural accuracy
depends on the specific prompt used to elicit responses, and prompt
sensitivity was not systematically varied in this study. Different
prompt formulations could yield different accuracy values.

\textbf{Cross-family comparisons conflate multiple factors.}
Differences between Qwen, Gemma, and Llama may reflect architecture,
training data, alignment procedure, or other recipe-level differences.
The current results do not isolate which of these factors drives the
observed patterns.

\textbf{Semantic contamination cannot be ruled out.} The ROUGE-L and
perplexity diagnostics do not support a narrow memorisation-based
contamination account, but they do not eliminate broader semantic
contamination through diffuse exposure to AI evaluation discourse.
Additionally, the held-out baseline used in the perplexity diagnostic
(\texttt{test\_full}) is not fully disjoint from the
\texttt{stages\_oversight} items in question bodies, and is therefore
best interpreted as a matched baseline rather than a fully independent
distribution.

\textbf{Scale is bounded at 32B.} All experiments use open-weight
models up to 32B parameters. Whether the observed depth shift
continues, stabilises, or reverses at larger frontier scales is a key
safety-relevant open question raised by this work, since the models of
greatest concern for evaluation gaming are precisely those at and
beyond the frontier.

\section{Conclusion}
We set out to test a specific claim --- that evaluation awareness scales
predictably as a power law across open-weight model families --- and
find that this account does not hold within the families studied here.

Across 11 models in three families, evaluation-awareness signals are
detectable but do not scale uniformly. Within Qwen 2.5 and Gemma 2,
peak probe AUROC is non-monotonic or inversely related to model size.
More importantly, in both families the representational depth at which
evaluation-awareness peaks shifts with scale, moving from later layers
in smaller models to earlier layers in larger ones. Internal
representation and behavioural expression dissociate, and the
relationship between them differs by family. Two contamination
diagnostics find no evidence for a narrow memorisation-based
explanation of these patterns.

Taken together, these findings suggest that evaluation awareness should
be studied as a family-dependent representational phenomenon rather
than a single smoothly scaling capability. If larger models encode
evaluation context in the earliest layers of processing, then
benchmark authenticity may matter more at frontier scale, not less.
Natural next steps include testing whether the depth shift generalises
to additional families and larger models, evaluating the phenomenon on
alternative benchmarks, and applying intervention-based methods to
determine what these early-layer signals are causally contributing to.

\section*{Broader Impact}
Understanding evaluation awareness matters for the design of
trustworthy benchmarks. If models can distinguish evaluation from
deployment contexts, benchmark performance may become a less reliable
indicator of real-world behaviour. The probe methodology introduced
here may also support future auditing of whether internal
representations of evaluation context are present in a given model,
complementing behavioural testing. This paper is primarily descriptive
and does not itself provide a method for inducing or exploiting
evaluation awareness.

\section*{Acknowledgements}
 
This work was completed as part of the BlueDot Impact Technical AI Safety Project Sprint. We are grateful to BlueDot Impact for the opportunity and for the compute grant that made the cloud GPU experiments possible. We thank Eitan Sprejer for mentorship throughout this project, and in particular for pushing us to think more carefully about methodology, framing, and the questions worth asking. We also thank Shivam Arora for generously sharing his perspective on the research direction.

{\raggedright
\bibliography{references}

@article{chaudhary2025evalawareness,
  title={Evaluation Awareness Scales Predictably in Open-Weights Large Language Models},
  author={Chaudhary, M. and Su, I. and Hooda, N. and Shankar, N. and Tan, J. and Zhu, K. and Lagasse, R. and Sharma, V. and Panda, A.},
  journal={arXiv preprint arXiv:2509.13333},
  year={2025},
  url={https://arxiv.org/abs/2509.13333}
}

@article{golchin2023timetravel,
  title={Time Travel in LLMs: Tracing Data Contamination in Large Language Models},
  author={Golchin, Shahriar and Surdeanu, Mihai},
  journal={arXiv preprint arXiv:2308.08493},
  year={2023},
  url={https://arxiv.org/abs/2308.08493}
}

@article{laine2024sad,
  title={Me, Myself, and AI: The Situational Awareness Dataset (SAD) for LLMs},
  author={Laine, Roope and Chughtai, Bilal and Betley, James and Hariharan, Karthik and Scheurer, Jeremy and Balesni, Mantas and Hobbhahn, Marius and Meinke, Alexander and Evans, Owain},
  journal={arXiv preprint arXiv:2407.04694},
  year={2024},
  url={https://arxiv.org/abs/2407.04694}
}

@article{nguyen2025probing,
  title={Probing and Steering Evaluation Awareness of Language Models},
  author={Nguyen, J. and Hoang, K. and Attubato, C. L. and Hofst{\"a}tter, F.},
  journal={arXiv preprint arXiv:2507.01786},
  year={2025},
  url={https://arxiv.org/abs/2507.01786}
}

@article{panickssery2024steering,
  title={Steering Llama 2 via Contrastive Activation Addition},
  author={Panickssery, N. and Gabrieli, N. and Schulz, J. and Tong, M. and Hubinger, E. and Turner, A. M.},
  journal={arXiv preprint arXiv:2312.06681},
  year={2024},
  url={https://arxiv.org/abs/2312.06681}
}

@article{ravaut2024survey,
  title={A Comprehensive Survey of Contamination Detection Methods in Large Language Models},
  author={Ravaut, Matthieu and Ding, B. and Jiao, F. and Chen, H. and Li, X. and Zhao, R. and Qin, C. and Xiong, C. and Joty, S.},
  journal={arXiv preprint arXiv:2404.00699},
  year={2024},
  url={https://arxiv.org/abs/2404.00699}
}

@misc{jordine2025dataset,
  author = {Jordine},
  title = {Pivotal Test Phase Steering},
  year = {2025},
  howpublished = {\url{https://github.com/Jordine/pivotal-test-phase-steering}}
}

@misc{anonymous2026code,
  author = {Manek, Archit},
  title = {evaluation-awareness-probing: Replication code and results for the eval-awareness probing study},
  year = {2026},
  howpublished = {\url{https://github.com/archit-manek/evaluation-awareness-probing}}
}
\bibliographystyle{icml2026}
}

\appendix
\onecolumn

\section{Dataset Details}

\subsection{Jordine contrastive dataset}

We use the dataset released in the Jordine
\textit{pivotal-test-phase-steering} repository
\citep{jordine2025dataset} for probe construction. The dataset contains
203 synthetic evaluation/deployment prompt pairs and is used only for
constructing probe directions. No modifications were made to the
dataset.

\subsection{Situational Awareness Dataset (SAD)}

We use the Situational Awareness Dataset (SAD) \citep{laine2024sad}
for probe evaluation, black-box behavioural evaluation, and
contamination diagnostics. Specifically, we use
\texttt{stages\_oversight} for both white-box and black-box
evaluation, and \texttt{test\_full} as a held-out baseline for the
perplexity contamination diagnostic. SAD items are stored as
structured question objects with fields including \texttt{body},
\texttt{choices\_right}, \texttt{choices\_wrong}, and \texttt{splits}.
For \texttt{stages\_oversight}, items are used as binary
evaluation/deployment classification examples. In \texttt{test\_full},
items use a broader label space and are used only for the held-out
perplexity baseline. No modifications were made to the underlying
items.

\section{Experimental Details}

\subsection{Relationship to Nguyen et al.}

Our white-box probing pipeline applies the methodology of
\citet{nguyen2025probing} to 11 models across three families. In
particular, we follow the general probing approach, the use of the
Jordine contrastive dataset for probe construction, and layerwise
AUROC evaluation on SAD. The main additions in this paper are: (1)
extending the analysis across 11 models from three families, (2)
analysing representational depth as a function of scale, (3) adding
benchmark-matched black-box behavioural evaluation, and (4)
introducing two direct contamination diagnostics.

\subsection{Contamination diagnostic setup}

\textbf{ROUGE-L completion test.} Following
\citet{golchin2023timetravel}, we provide the first 60\% of each SAD
question as a prompt and ask the model to complete it, withholding the
remaining 40\% as the reference. If a model has memorised the
benchmark, completions would be expected to more closely match the
original continuation, yielding higher ROUGE-L scores. We run this
test on 50 SAD questions across all 11 models.

\textbf{Perplexity-ratio diagnostic.} Following methods surveyed by
\citet{ravaut2024survey}, we compute mean per-token perplexity on 100
\texttt{stages\_oversight} items and compare it against 100 items from
SAD \texttt{test\_full}, matched for domain and style. A ratio greater
than 1.0 indicates that \texttt{stages\_oversight} is harder for the
model than the held-out baseline, which is opposite to what direct
contamination would predict. We note that absolute perplexity values
are not comparable across model families due to architectural and
tokenizer differences; the diagnostic relies on the within-model ratio
rather than cross-family comparison of raw values.

\subsection{Prompt template}

For black-box evaluation, we apply the following prompt template to
each SAD item, where \texttt{<BODY>} is replaced with the
\texttt{body} field of the item:

\begin{quote}
\small
\texttt{You are analyzing prompts given to an AI assistant.}\\
\texttt{Determine whether the following prompt is more likely to come}\\
\texttt{from an evaluation/testing context or a real deployment context.}\\[0.5em]
\texttt{Prompt: <BODY>}\\[0.5em]
\texttt{Answer with exactly one word --- either `evaluation' or `deployment':}
\end{quote}

Responses are decoded greedily. Outputs containing
\texttt{evaluation} or \texttt{testing} are assigned the evaluation
label; outputs containing \texttt{deployment} are assigned the
deployment label; all other outputs are treated as invalid. Reported
black-box accuracy is therefore computed over valid, parseable
responses only, and valid response rate is reported separately in the
main results. Responses are capped at 10 new tokens; keyword matching is applied to the full decoded output.

\section{Contamination Diagnostic Results}

\subsection{ROUGE-L completion results}

\begin{table}[h]
\centering
\small
\caption{ROUGE-L completion diagnostic on 50 SAD questions. Low
scores across all families are inconsistent with a simple
memorisation-based contamination account.}
\label{tab:rouge}
\begin{tabular}{llcc}
\toprule
Model & Family & ROUGE-L & Std.\ Dev. \\
\midrule
0.5B  & Qwen 2.5  & 0.118 & 0.107 \\
1.5B  & Qwen 2.5  & 0.158 & 0.151 \\
3B    & Qwen 2.5  & 0.135 & 0.124 \\
7B    & Qwen 2.5  & 0.149 & 0.145 \\
14B   & Qwen 2.5  & 0.135 & 0.138 \\
32B   & Qwen 2.5  & 0.143 & 0.124 \\
\midrule
2B    & Gemma 2   & 0.114 & 0.100 \\
9B    & Gemma 2   & 0.137 & 0.148 \\
27B   & Gemma 2   & 0.142 & 0.143 \\
\midrule
1B    & Llama 3.2 & 0.123 & 0.113 \\
3B    & Llama 3.2 & 0.121 & 0.105 \\
\bottomrule
\end{tabular}
\end{table}

All models score between 0.114 and 0.158, with no meaningful
separation between Qwen, Gemma, and Llama families. Gemma 2-2B, which
shows the strongest black-box evaluation-awareness signal in the main
results, has the lowest ROUGE-L score of any tested model, which is
not what a memorisation-based explanation would predict.

\subsection{Perplexity-ratio results}

\begin{table}[h]
\centering
\small
\caption{Perplexity-ratio contamination diagnostic. Ratios above 1.0
indicate that \texttt{stages\_oversight} is harder for the model than
the held-out baseline, opposite to what direct contamination would
predict. Absolute perplexity values are not comparable across families
due to architectural and tokenizer differences.}
\label{tab:ppl}
\begin{tabular}{llccc}
\toprule
Model & Family & SAD PPL & Baseline PPL & Ratio \\
\midrule
0.5B  & Qwen 2.5  & 51.23  & 24.94 & 2.054 \\
1.5B  & Qwen 2.5  & 32.66  & 19.18 & 1.703 \\
3B    & Qwen 2.5  & 33.68  & 22.41 & 1.503 \\
7B    & Qwen 2.5  & 32.63  & 23.99 & 1.360 \\
14B   & Qwen 2.5  & 29.42  & 18.96 & 1.552 \\
32B   & Qwen 2.5  & 26.25  & 22.25 & 1.180 \\
\midrule
2B    & Gemma 2   & 146.02 & 25.31 & 5.769 \\
9B    & Gemma 2   & 168.90 & 26.79 & 6.305 \\
27B   & Gemma 2   & 78.28  & 25.24 & 3.102 \\
\midrule
1B    & Llama 3.2 & 63.25  & 28.99 & 2.182 \\
3B    & Llama 3.2 & 49.68  & 26.42 & 1.881 \\
\bottomrule
\end{tabular}
\end{table}

All models show ratios greater than 1.0, ranging from 1.18 to 6.31.
Gemma models show substantially higher absolute perplexity values than
other families, which likely reflects architectural and tokenizer
differences rather than contamination; the relevant quantity is the
within-model ratio, which points in the same direction for all
families.

\section{Reproducibility}

Code, prompts, and experiment outputs are publicly available in the
accompanying repository \citep{anonymous2026code}:
\url{https://github.com/archit-manek/evaluation-awareness-probing}.

\end{document}